\documentclass[conference]{IEEEtran}
\IEEEoverridecommandlockouts

\usepackage{cite}
\usepackage{amsmath,amssymb,amsfonts}
\usepackage{algorithmic}
\usepackage{graphicx}
\usepackage{textcomp}
\usepackage{xcolor}
\usepackage{booktabs}
\def\BibTeX{{\rm B\kern-.05em{\sc i\kern-.025em b}\kern-.08em
		T\kern-.1667em\lower.7ex\hbox{E}\kern-.125emX}}
	
	\usepackage{multirow}
	\newcommand{\citep}{\cite}
	\usepackage{amssymb}
	\usepackage{pifont}

\def\BibTeX{{\rm B\kern-.05em{\sc i\kern-.025em b}\kern-.08em
		T\kern-.1667em\lower.7ex\hbox{E}\kern-.125emX}}

\title{\LARGE \bf
	DURableVS: Data-efficient Unsupervised Recalibrating Visual Servoing via online learning in a structured generative model
}

\author{Nishad Gothoskar$^{1}$\qquad Miguel L\'{a}zaro-Gredilla$^{2}$\\Yasemin Bekiroglu$^{2}$\qquad Abhishek Agarwal$^{2}$\\Joshua B. Tenenbaum$^{1}$\qquad Vikash K. Mansinghka$^{1}$\qquad Dileep George$^{2}$\\ \\
\thanks{$^{1}$Massachusetts Institute of Technology $^{2}$Vicarious AI}
}%

\begin{document}
\maketitle


\begin{abstract}
Visual servoing enables robotic systems to perform accurate closed-loop control, which is required in many applications. However, existing methods either require precise calibration of the robot kinematic model and cameras or use neural architectures that require large amounts of data to train. In this work, we present a method for unsupervised learning of visual servoing that does not require any prior calibration and is extremely data-efficient. Our key insight is that visual servoing does not depend on identifying the veridical kinematic and camera parameters, but instead only on an accurate generative model of image feature observations from the joint positions of the robot. We demonstrate that with our model architecture and learning algorithm, we can consistently learn accurate models from less than 50 training samples (which amounts to less than 1 min of unsupervised data collection), and that such data-efficient learning is not possible with standard neural architectures. Further, we show that by using the generative model in the loop and learning online, we can enable a robotic system to recover from calibration errors and to detect and quickly adapt to possibly unexpected changes in the robot-camera system (e.g. bumped camera, new objects).
\end{abstract}	



\section{Introduction}

In robotics, techniques that use visual feedback to guide robot control are referred to as visual servoing \cite{kragic2002survey,chaumette2016visual, siciliano2016springer}. But while visual servoing enables closed-loop control, it often relies on calibration of the robot-camera system \cite{tsai1980review}. In particular, it requires precise calibration of the cameras observing the robot and the kinematic model of the robot itself \cite{marchand2005visp}. Calibration procedures may require human input or precisely manufactured devices and are often tedious to perform. Recent work has attempt to circumvent the need for calibration by instead learning visual servoing, often using neural network architectures \cite{levine2018learning, sadeghi2018sim2real}. However, these approaches require large amounts of data and time to train in order to achieve good performance. Further, they cannot quickly adapt to changes in the robot-camera system (e.g. bumped camera, new objects) without retraining.


\begin{figure}[t]
	\centering\includegraphics[width=\linewidth]{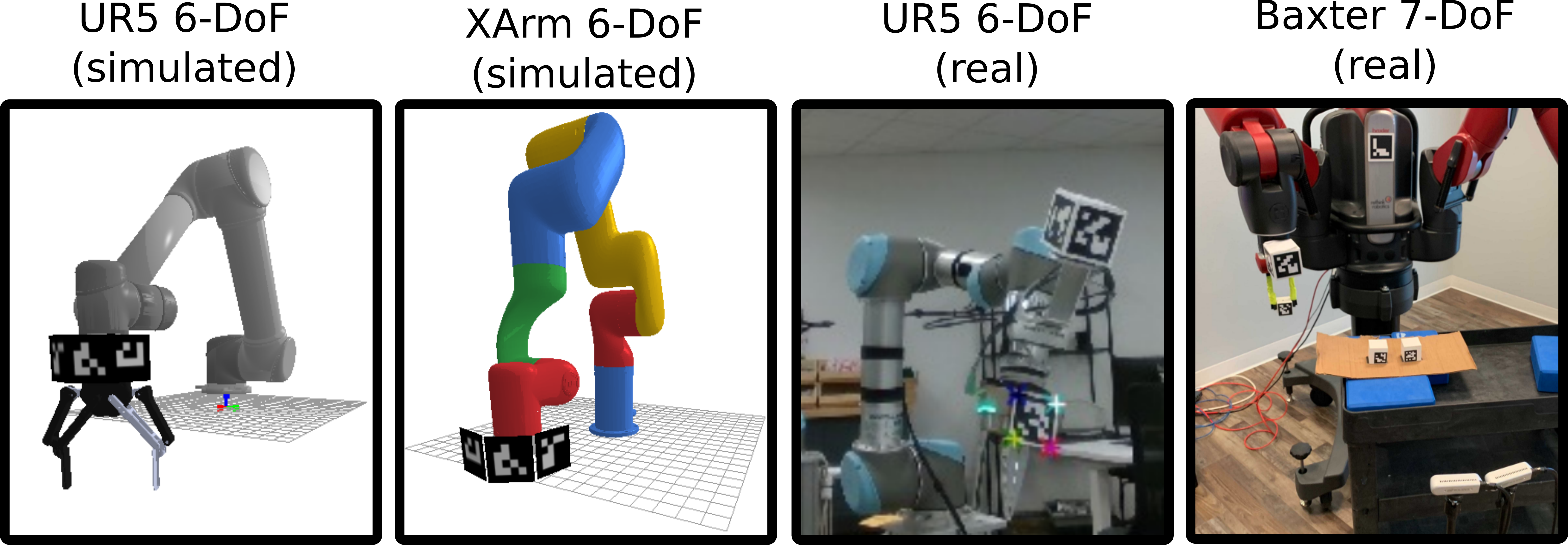}
	\caption{Evaluation environments. We evaluated DurableVS in 4 different environments: two simulated environments with a UR5 (6-DoF) and Xarm (6-DoF) and two real environments with a UR5 and a Baxter (7-DoF).}
	\label{fig:setup}
\end{figure}

\begin{table}[]
	\centering
	       \fontsize{8.5pt}{8.5pt}\selectfont
      \renewcommand{\arraystretch}{1.5}
	\caption{DurableVS versus VISP \cite{marchand2005visp} versus neural method}
	\begin{tabular}{|c|c|c|c|}
		\hline
		\textbf{Method} & \textbf{Data-Efficient} & \textbf{Unsupervised} &  \textbf{Re-calibrating}  \\ \hline
		VISP \cite{marchand2005visp} & \ding{52} & requires calibration & \\ \hline
		Neural &  & \ding{52} &  \\ \hline
		Ours & \ding{52} & \ding{52} & \ding{52}\\ \hline
	\end{tabular}

\end{table}

\begin{figure*}[]	
	\centering
	\includegraphics[width=0.9\textwidth]{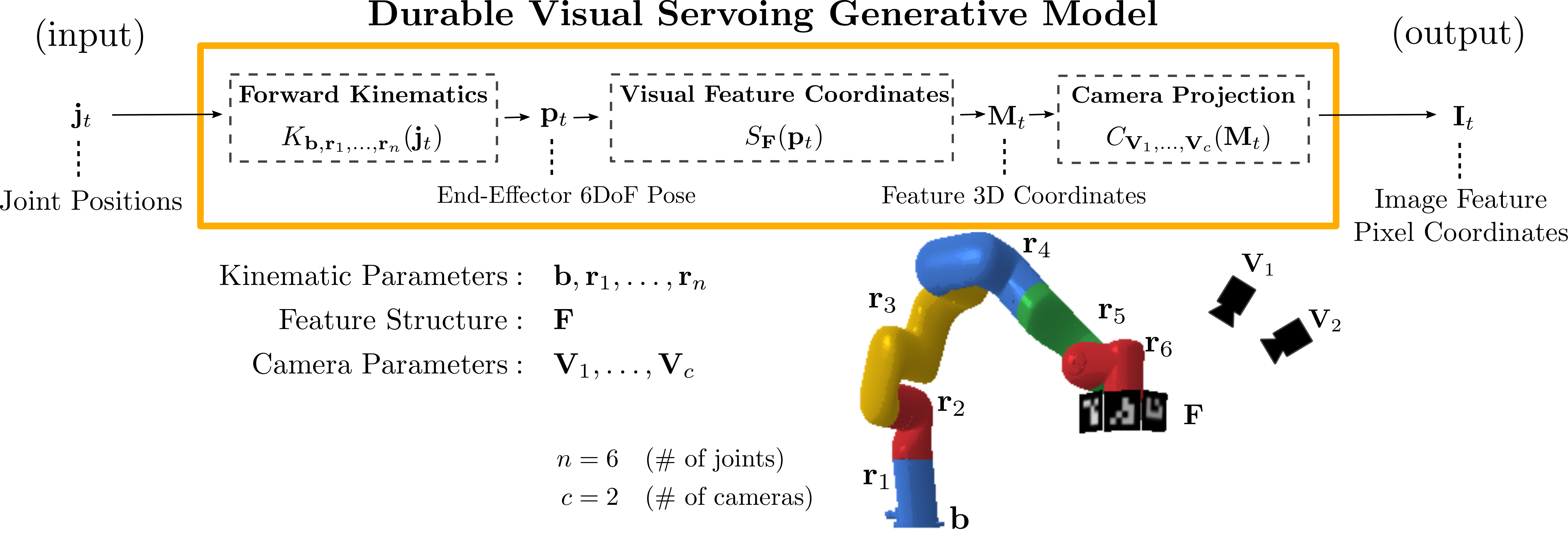}
	\caption{We illustrate the 3 modules that are composed to define our generative model from joint positions $\mathbf{j}_t$ to image observations $\mathbf{I}_t$, and the parameters of those modules.  According to the kinematic structure of the robot, the joint positions $\mathbf{j}_t$ will place the robot's end-effector at a certain 6D Cartesian pose $\mathbf{p}_t$. Given the robot end-effector's pose and the relative coordinates of the visual features in the end-effector frame, we can compute the 3D coordinates $\mathbf{M}_t$ of the features in the world frame. Finally, the features are observed through the cameras and appear at pixel coordinates $\mathbf{I}_t$ in the images.}
	\label{fig:model}
\end{figure*}

In this paper, we present a learning-based approach to visual servoing that, rather than employing neural networks, uses structured generative models. We propose a differentiable model architecture and optimization procedure for learning the generative model's parameters from data. While our architecture resembles standard kinematic and camera modeling, a key insight of our work is that we can perform visual servoing without identifying the veridical (real world) kinematic and camera parameters, but rather just using an accurate generative model of image feature coordinates from joint positions.
Our experimental results demonstrate that we can learn accurate models extremely data-efficiently, from \textless 50 training samples (\textless 1 min of unsupervised data collection), and that such data-efficient learning is not possible with standard neural architectures. Further, since our model is fast enough to run in the control loop and learning can be done online, our method can be used to detect and adapt to unexpected changes in the robot-camera system.

In the remainder of this paper, we introduce our model architecture, learning algorithm, and inference procedure. Then we present experimental results highlighting our approach's accuracy, data-efficiency, speed, and its ability to quickly recover from changes to the robot-camera setup via online learning. We conclude by discussing related work in vision-based control, limitations, and future work.
\section{DurableVS Model}
\label{sec:model}

Our architecture is composed of three modules: (1) Forward Kinematics, (2) Visual Feature Structure, and (3) Camera Projection. The \textit{Forward Kinematics} module takes as input the joint positions $\mathbf{j}_t$ and predicts the 6DoF Cartesian pose $\mathbf{p}_t$ of the end-effector. The \textit{Visual Feature Structure} module takes as input the 6DoF Cartesian pose $\mathbf{p}_t$ of the end-effector and predicts the 3D coordinates $\mathbf{M}_t$ of the visual features that are rigidly attached to the end-effector. The \textit{Camera Projection} module takes as input the 3D coordinates $\mathbf{M}_t$ of the visual features and predicts the pixel coordinates $\mathbf{I}_t$ at which the visual features appear in the observed images. We compose these 3 modules, in sequence, to get a model that, given the joint positions $\mathbf{j}_t$ as input, predicts the pixel coordinates $\mathbf{I}_t$. This architecture parallels the causal generative process by which placing a robot at certain joint positions results in visual features attached to the end-effector being observed at certain locations in the images. (See Figure \ref{fig:model}). We now discuss the specific parametrization of each of the 3 modules:


\textbf{Forward Kinematics} We parametrize the Forward Kinematics module using the Denavit–Hartenberg (DH) convention \cite{hartenberg1964kinematic}. For each of the $n$ links of an $n$-DoF robot, there are 4 DH parameters: $\omega$ - joint angle offset; $d$ - link offset; $a$ - link length; $\alpha$ - link twist. We collect these parameters per link into vectors $\mathbf{r}_1,\mathbf{r}_2,\dots,\mathbf{r}_n$. The link parameters $\mathbf{r}_i$ and the current joint angle $\mathbf{j}_t[i]$ together define a relative transformation $L_{\mathbf{r_i }}(\mathbf{j}_t[i])$ between consecutive links. Composing these relative transformation for all links in the kinematic chain gives the pose of the end-effector in the robot's base frame. We additionally parametrize the 6DoF pose of the base frame as $\mathbf{b}$. The Forward Kinematics $K$ takes as input the joint positions $\mathbf{j}_t$ and outputs the 6DoF Cartesian end-effector pose $\mathbf{p}_t$:

\begin{equation}
\mathbf{p}_t = K_{\mathbf{b}, \mathbf{r}_1,\dots,\mathbf{r}_n}(\mathbf{j}_t) = \mathbf{b} \cdot \prod_{i = 1}^{n} L_{\mathbf{r}_i} (\mathbf{j}_t[i])
\end{equation}
The parameters that must be learned are: $\mathbf{b}, \mathbf{r}_1,\dots,\mathbf{r}_n$. We refer to them collectively as $\mathbf{R}$.


\textbf{Visual Feature Structure}
We assume that we can track visual features that are rigidly attached to the robot's end-effector. Thus, these features can be described by their 3D (relative) coordinates $\mathbf{F}$ in the end-effector's reference frame. The Visual Feature Structure module $S$ takes as input the 6DoF Cartesian pose $\mathbf{p}_t$ of the end-effector in the world frame and outputs the 3D coordinates $\mathbf{M}_t$ of the visual features in the world frame, done via a homogeneous transformation:

\begin{equation}
\mathbf{M}_t = S_{\mathbf{F}}(\mathbf{p}_t) = \mathbf{p}_t \cdot  \mathbf{F}^\top
\end{equation}
The parameter that must be learned is: $\mathbf{F}$.

\textbf{Camera Projection}
We model cameras using a pinhole camera model whose parameters are the extrinsics matrix $\mathbf{E}$, which defines the relative pose of the camera in the world frame, and the intrinsic matrix $\mathbf{K}$, which governs how 3D points in the camera frame are projected onto the image plane via focal lengths and principal points \cite{szeliski2010computer}. Each of the $c$ cameras has its own parameters and we use $\mathbf{K}_i$ and $\mathbf{E}_i$ to denote the intrinsics and extrinsics of camera $i$, respectively. The Camera Projection $C$ takes as input the 3D homogeneous coordinates $\mathbf{M}_t$ of the features and outputs the 2D coordinates of those features in the image $\mathbf{I}^{(i)}_t$ as observed in camera $i$:

\begin{equation}
\mathbf{I}^{(i)}_t = C_{\mathbf{K}_i,\mathbf{E}_i}(\mathbf{M}_t) = \mathbf{K} \cdot \mathbf{E} \cdot \mathbf{M}_t
\end{equation}
The parameters that must be learned are: $\mathbf{K}_i$ and $\mathbf{E}_i$ for each camera $i=1,\dots,c$ where $c$ is the number of cameras. We collect the intrinsics and extrinsics per camera and refer to them as $\mathbf{V}_i$ for camera $i$. To simplify notation, when we call $C$ on the input $\mathbf{M}_t$ with all camera parameters $\mathbf{V}_1,\dots,\mathbf{V}_c$, we assume the output is the predicted image features $\mathbf{I}^{(1)}_t,\dots,\mathbf{I}^{(c)}_t$ in all $c$ cameras, which we collect and refer to as $\mathbf{I}_t$.

\textbf{Summary} Our architecture composes the 3 modules described above:

\begin{alignat*}{2}
\text{Forward Kinematics:}& \quad \mathbf{p}_t &&= K_{\mathbf{R}}(\mathbf{j}_t)\\
\text{Visual Feature Structure:}& \quad \mathbf{M}_t &&= S_{\mathbf{F}}(\mathbf{p}_t)\\
\text{Camera Projection:}& \quad \mathbf{I}_t &&= C_{\mathbf{V_1},\dots,\mathbf{V_c}}(\mathbf{M}_t)\\
\textbf{Full Architecture:}& \quad \mathbf{I}_t &&= C_{\mathbf{V_1},\dots,\mathbf{V_c}} (S_{\mathbf{F}}(  K_{\mathbf{R}}(\mathbf{j}_t)  )  )
\end{alignat*}

Importantly, each of the above modules is differentiable and thus the full architecture is differentiable.

\section{Learning}
\label{sec:learning}

We learn the parameters of our model from training data consisting of pairs $(\mathbf{j}_t, \mathbf{I}_t)$ of joint positions and corresponding image feature coordinates. This data can be collected in a completely unsupervised manner by sending random actions to the robot and capturing images from the cameras. Given this data, we learn the parameters of the model via stochastic gradient-based optimization with L-BFGS \cite{liu1989limited} optimizing the reconstruction error between the predictions of the model (given  $\mathbf{j}_t$ as input) and the true observations $\mathbf{I}_t$.

While the ground truth veridical parameters (i.e. the true robot kinematic model, feature structure, and camera intrinsics and extrinsics) represent one solution to this optimization problem, it is not the only solution. Consider parameter settings corresponding to scaled up, scaled down, rotated, and translated version of the real world. These models would be equally accurate at predicting image observations from joint positions, even though the model's latent 3D representation may not necessarily coincide with the real world. A key insight of our work is that these models can be used for visual servoing and can be learned unsupervised.

In the remainder of this section, we describe the 3 subroutines of our learning algorithm:

\subsection{Camera and Feature Structure Learning}
\label{sec:camandfeature}

We first optimize the camera parameters $\mathbf{V}_{1:c}$ (for all $c$ cameras) and feature coordinate parameters $\mathbf{F}$. In addition to optimizing these parameters, we also optimize the end-effector poses $\mathbf{p}_{1:T}$, which are a variable in the model. In this first stage of the optimization, we are considering a truncated model from poses $\mathbf{p}_t$ to pixel coordinates $\mathbf{I}_t$, and ignoring the joint positions and kinematics. The function $D_{\textrm{pixel}}$ is a quadratic error between the observed and generated pixel coordinates. The objective of the following optimization is to find a setting of the parameters ($\mathbf{V}_{1:c}$,$\mathbf{F}$) and poses ($\mathbf{p}_{1:T}$) that produces pixel coordinates consistent with the actual observed coordinates $\mathbf{I}_{t}$ (for all $c$ cameras and $T$ time steps):
\begin{equation}
\label{eq:losscamstruct}
\begin{gathered}
\mathbf{F}^*, \ \mathbf{V}_{1:c}^*, \ \mathbf{p}_{1:T}^* = \\ \underset{\mathbf{F}, \mathbf{V}_{1:c}, \mathbf{p}_{1:T}}{\text{argmin}} \sum_{t=1}^{T} \sum_{i=1}^{c} D_{\text{pixel}}\big(  C_{\mathbf{V}_i}(  S_\mathbf{F}( \ \mathbf{p}_t  )  ), \mathbf{I}_{t}^{(i)} \big)
\end{gathered}
\end{equation}
$\mathbf{I}_{t}^{(i)}$ are the feature detections at time $t$ in camera $i$. This optimization is prone to convergence to sub-optimal local minima and it is important to provide good initialization for $\mathbf{p}_{1:T}$ and $\mathbf{V}_{1:c}$. We compute initializations by triangulating features observed in multiple cameras and estimating the camera baselines i.e the transforms between pairs of cameras.

\subsection{Kinematic Learning}
\label{sec:kinlearning}
Next, we optimize the kinematic parameters $\mathbf{R}$. We use the observed joint positions $\mathbf{j}_{1:T}$ and the Cartesian end-effector poses $\mathbf{p}_{1:T}$ output from the previous subroutine. The function $D_{\text{pose}}(\mathbf{A}, \mathbf{G}) = \| \mathbf{A} - \mathbf{G} \|_F$ is the Frobenius norm between pose matrices $\mathbf{A}$ and $\mathbf{G}$. The objective of the following optimization is to find a setting of the kinematic parameters $\mathbf{R}$ that produces Cartesian poses consistent with $\mathbf{p}_{1:T}^*$.

\begin{equation}
\label{eq:losskin}
\begin{gathered}
\mathbf{R}^* = \underset{\mathbf{R}}{\text{argmin}} \ \sum_{t=1}^{T} D_{\text{pose}}(K_{\mathbf{R}}(\mathbf{j}_t),\mathbf{p}_t^*) 
\end{gathered}
\end{equation}

Note, we are using $\mathbf{p}_{1:T}^*$, the output of the previous subroutine, as a target in this optimization. Since these were learned without considering kinematic structure, they may be inaccurate and therefore, the full model optimization (described in the next section), which simultaneously optimizes all model parameters, is needed to correct these errors.

\subsection{Full Model Learning}
\label{sec:e2elearning}
Finally, we optimize all model parameters simultaneously. This optimization is very similar to that of the \textit{Camera and Structure} learning (Section \ref{sec:camandfeature}). The error is also between the generated and observed pixel coordinates $\mathbf{I}_{t}$. However instead of using the truncated model (from $\mathbf{p}_t$ to $\mathbf{I}_t$) we consider the full model (from $\mathbf{j}_t$ to $\mathbf{I}_t$).

\begin{equation}
\label{eq:losse2e}
\begin{gathered}
\mathbf{R}^f, \mathbf{F}^f, \mathbf{V}_{1:c}^f = \\
\underset{\mathbf{R}, \mathbf{F}, \mathbf{V}_{1:c}}{\text{argmin}} \sum_{t=1}^{T} \sum_{i=1}^{c} D_{\text{pixel}}(C_{\mathbf{V}_i}(S_\mathbf{F}(K_{\mathbf{R}}(\mathbf{j}_t))),\mathbf{I}_{t}^{(i)})
\end{gathered}
\end{equation}

The outputs of the previous two subroutines ($\mathbf{R}^*$, $\mathbf{F}^*$, $\mathbf{V}^*_{1:c}$) serve as the initialization for this optimization. We found that all 3 subroutines of this optimization are needed to ensure consistent convergence. Table \ref{tab:data} compares to DVS-OnlyFull, an ablation of our method that does not use the other 2 subroutines for learning, and we find that it sometimes fails to learn accurate models, especially as we reduce the amount of training data.
%

\section{Inference}
\label{sec:inference}

After learning the model parameters, the model can predict the coordinates $\mathbf{I}_t$ of the image features given the robot's joint positions $\mathbf{j}_t$. However, in order to use this model for visual servoing, we need to be able to find the joint positions $\mathbf{\hat{j}}_t$ such that the image features will be located at some desired target locations $\mathbf{\hat{I}}_t$ in the image. We use an inference-via-optimization approach to infer $\mathbf{\hat{j}}_t$:

\begin{equation}
\label{eq:inf_optimization}
\begin{gathered}
\mathbf{\hat{j}}_t = \underset{\mathbf{j}_t}{\text{argmin}} \ \  D_{\text{pixel}}\big(f(\mathbf{j}_t), \ \mathbf{\hat{I}}_t)
\end{gathered}
\end{equation}
where $f$ is our learned model. 

We can then use the above inference to control the robot. We first infer the current joint positions $\mathbf{j}_{t}$ from the current observations $\mathbf{I}_{t}$. Then, we select desired locations of the image features $\mathbf{\hat{I}}_t$ and infer the corresponding joint positions $\mathbf{\hat{j}}_t$. Finally, we command the robot to move by $\mathbf{\hat{j}}_t - \mathbf{j}_{t}$ to match those desired image feature coordinates. This optimization is implicitly solving optimizations corresponding to Inverse Kinematics (IK) \cite{beeson2015trac} and Perspective-n-Point \cite{fischler1981random} problems.

\section{Experiments}
\label{sec:results}

We conducted several experiments that demonstrate the accuracy, data-efficiency, speed, and flexibility of our approach. We first present a quantitative evaluation comparing our method, ablations of our method, and neural baselines. The results show that our approach learns very accurate models more data-efficiently than neural baselines. Then, we show the speed of our generative model, demonstrating that our approach is usable for real-time applications. Next, we show that our method can detect and quickly adapt to changes in the robot-camera setup via online learning. Finally, we evaluate our method's servoing accuracy, showing comparable accuracy to VISP \cite{marchand2005visp}, a standard calibrated method for visual servoing.

\subsection{Data-Efficiency} 

We first demonstrate the data-efficiency of our method on real and simulated robot environments. Table \ref{tab:data} shows a quantitative evaluation of our method compared to neural network baselines and ablations of our method. Each method is trained on the indicated number of training samples (pairs of $(\mathbf{j}_t, \mathbf{I}_t)$) and tested on a held-out data set of 100 samples. We report the average error between the predicted and observed pixel coordinates (for visual features that were observed).  We compare with 4 neural architecture baselines. \textbf{NN1}: Single hidden layer (200 units), Tanh activation.  \textbf{NN2}: Single hidden layer (200 units), ReLU activation. \textbf{NN3}: Two hidden layers (100, 50 units), Tanh, ReLU activations. \textbf{NN4}: Two hidden layers (100, 50 units), ReLU, ReLU activations. We also compare with 2 ablations of our model. \textbf{DVS-NoFull}: \textit{Camera and Feature Structure Learning} (Section \ref{sec:camandfeature}) and \textit{Kinematic Parameter Learning} are used but not \textit{Full Model Learning} (Section \ref{sec:e2elearning}). \textbf{DVS-OnlyFull}: Only \textit{Full Model Learning} (Section \ref{sec:e2elearning})  is used. Our full architecture is denoted \textbf{DVS}, which uses all 3 subroutines for learning the model parameters. We find that our method consistently outperforms the baselines in all environments. The neural networks are unable to match the performance of our model given the same amount of data.

This is due in part to the concise parametrization of our model, which prevents overfitting the limited training data. The neural architecture lacks the structure of our model and is significantly overparametrized. Note that for an $n$ degree of freedom (DOF) robot with $m$ tracked features that are observed in $c$ cameras, our model's parameters are: $6$ for the robot base frame, $4n$ for DH parameters for all joints, $3m$ for 3D coordinates of each of the features, $6c$ for the camera extrinsics and $4c$ for camera intrinsics, totaling $6 + 4n + 3m + 10c$ parameters. In our experiments where we have a $6$DoF robot, $12$ features, and $2$ cameras, the total number of parameters is $86$. In comparison, a neural architecture with just $50$ hidden units already has $2798$ parameters.

We also compare with two ablations of our method, DVS-NoFull and DVS-OnlyFull, which don't use the full learning algorithm but only some subset of the 3 subroutines. DVS-OnlyFull, the ablation that only uses the \textit{Full Model Learning} subroutine (Section \ref{sec:e2elearning}), can sometimes match the performance of our full method DVS when 75 or 100 training samples are available. However, when only 50 samples are available the full method outperforms the ablations, illustrating the need for all 3 subroutines.

\begin{table*}[h!]
	\fontsize{8.5pt}{8.5pt}\selectfont
	\caption{Reconstruction error of learned models on each of the 4 environments, varying the training data set size
	}
	\centering\begin{tabular}{ll|lllllll}
		\toprule
		& &    \multicolumn{7}{c}{\textbf{Reconstruction Error (Test)}}
		\\
		\\
		& &\multicolumn{4}{c}{\textbf{Neural Baselines}}&
		\multicolumn{3}{c}{\textbf{DurableVS (DVS)}} \\
		\textbf{Environment} & \textbf{\# Samples} & NN1 & NN2 & NN3 & NN4 & DVS-NoFull & DVS-OnlyFull & DVS \\
		\midrule
		
		\multirow{3}{*}{UR5 (sim)} & 50 & 26.45 & 13.87 & 70.51 & 12.43  & 10.91 & 8.92 & \textbf{0.64} \\
		&  75 & 15.00 & 14.35 & 85.45 & 9.15  & 11.40 & 4.36 & \textbf{0.64} \\
		& 100 & 20.45 & 11.35 & 79.34 & 10.36  & 12.54 & 3.04 & \textbf{0.63} \\
		\midrule
		\multirow{3}{*}{Xarm (sim)} & 50 & 39.27 & 27.15 & 27.82 & 23.00  & 7.54 & 31.02 & \textbf{0.37} \\
		&  75 & 38.24 & 19.47 & 29.05 & 16.08  & 7.58 & 0.36 & \textbf{0.36} \\
		& 100 & 33.80 & 14.76 & 16.90 & 14.01  & 9.40 & \textbf{0.36} & \textbf{0.36} \\
		\midrule
		\multirow{3}{*}{UR5 (real)} & 50 & 60.28& 3.56& 29.61 & 4.48  & 6.46 & 4.64 & \textbf{0.36} \\
		&  75 & 60.27 & 23.00 & 28.99 & 3.96  & 9.28 & 4.25 & \textbf{0.37} \\
		&  100 & 60.27 & 3.62 & 59.69 & 5.58  & 6.53 & 4.84 & \textbf{0.36} \\
		\midrule
		
		\multirow{3}{*}{Baxter (real)} & 50 & 16.07& 13.54 & 16.25 & 14.76  & 7.87 & 12.17 & \textbf{2.40}\\
		& 75 & 12.61 & 11.02 & 14.48 & 8.88  & 12.36 & 2.21 & \textbf{2.20}\\
		& 100 & 13.83 & 9.57 & 11.31 & 7.93 & 7.58 & \textbf{2.01}&  \textbf{2.00}\\
		\bottomrule
	\end{tabular}
	\label{tab:data}
\end{table*}

\subsection{Speed}

For our architecture to be practically useful, it must be fast enough to be used for real-time servoing. In Table \ref{tab:speed}, we provide times for the forward, gradient, and inference computations in our model. Real-time servoing systems, such as VISP \cite{marchand2005visp}, use gradient-based servoing. Computing a gradient through our full model takes 16.1ms implying that our model supports 60+ FPS servoing, which would not be the bottleneck in most visual servoing systems given standard frame rates of cameras and frame rates of object/feature detectors that are often used for vision-based control.

\begin{table}
	\fontsize{7pt}{7pt}\selectfont
	\setlength{\tabcolsep}{9pt}
	\caption{Speed of queries in DurableVS}
	\centering\begin{tabular}{lll}
		\toprule
		\textbf{Query} & \textbf{Time (s)} & \textbf{FPS}\\
		\midrule
		Forward Kinematics & 0.0012 & 833\\
		Forward Camera/Feature & 0.0014 & 714\\
		Forward Full Model & 0.0025  & 400\\
		Gradient Full Model & 0.0161 & 62 \\
		Gradient Kinematics & 0.0126  & 79\\ 
		Gradient Camera/Feature & 0.0091 & 110 \\ 
		Infer Pose from Image & 0.0005  & 2000 \\
		Infer Joints from Pose & 0.1012  & 10\\ 
		Infer Joints from Image & 0.1025 & 10\\ 
		\hline
	\end{tabular}
	\label{tab:speed}
\end{table}

\subsection{Online Learning}

Most robotic systems assume that kinematic and camera calibrations will be performed once during setup and will remain fixed over the course of the robot's use. These system often do not monitor these calibrations, so if they were to change, this may only be detected in downstream task performance. For example, imagine a robot that uses an overhead camera to detect objects and plan grasps. If the camera's extrinsics were to drift by 1cm, the resulting grasps would be offset by 1cm, which would degrade the overall grasping performance. We demonstrate that by using our model, we can detect these changes to the system and adapt to them via online learning. We evaluate our method's ability to adapt to these changes by first learning the model parameters with an initial configuration of the robot-camera setup, then modifying the setup and measuring the model's reconstruction error as it sequentially receives new observations and updates the model parameters online. The reconstruction error is computed on a held-out data set collected in the new setup.

First, we consider modifications to the cameras (adding a new camera and moving an existing camera), which might happen accidentally if a camera is bumped or intentionally if another viewpoint is needed for a certain task. (We conducted this experiment on the UR5 real environment). Figure \ref{fig:graphs}(a) shows the reconstruction error as a function of the number of new samples available for the case where a third camera is added and the case where one of existing two cameras is moved. With just 2 new samples, the model is sufficiently accurate to begin servoing again and continues to improve its accuracy as more samples become available.


Next, we consider modifications to the structure of the visual features. Consider the setting where a robot grasps a peg and wants to insert it into a hole. To do this accurately, we would track and servo using visual features on the peg to align it to the hole. This would require learning the 3D structure of the new visual features on the peg. We evaluate our method's ability to learn new visual features, by attaching a novel marker to the robot's end-effector. (We conducted this experiment on the UR5 real environment). Figure \ref{fig:graphs}(b) shows the reconstruction error (of the new visual features) as a function of the number of new samples available. With 2 samples, the model is already very accurate. One explanation for this efficiency is to think that if we had two cameras properly calibrated, the new features could be triangulated to 3D, given pixel coordinates. With more views, the error in the triangulation reduces. Note, this triangulation is not done explicitly, only implicitly through the optimization. 

Finally, we may want to relearn the robot's kinematic model, for example, if a joint is damaged or if we modify the end-effector tool e.g. swap the gripper. (Since we could not modify the real robot kinematics, we conducted this experiment on the simulated UR5 robot by adding random noise to the robot's link lengths.) In Figure \ref{fig:graphs}(c), we show the reconstruction error as a function of the number of new samples available. Compared to re-learning camera and feature structure parameters, re-learning the kinematic parameters require more new samples. However, within approximately 25 samples the model has adapted and is very accurate again.

These results demonstrate that when there are unexpected changes to the robot-camera setup (e.g. bumped cameras, new visual features, damaged robot), modules of the generative model can be quickly relearned to adapt to those changes. But how do we detect these unexpected changes? We can detect these changes by comparing the expected image feature coordinates (as predicted by the learned generative model) with the actual observations. During normal operation, these should match very closely, however when there are unexpected changes to the system there will be discrepancies between the expectation and observation. We use this as a signal that model parameters need to be updated.

\subsection{Servoing Accuracy}

\begin{figure*}
	\centering
	\includegraphics[width=0.94\linewidth]{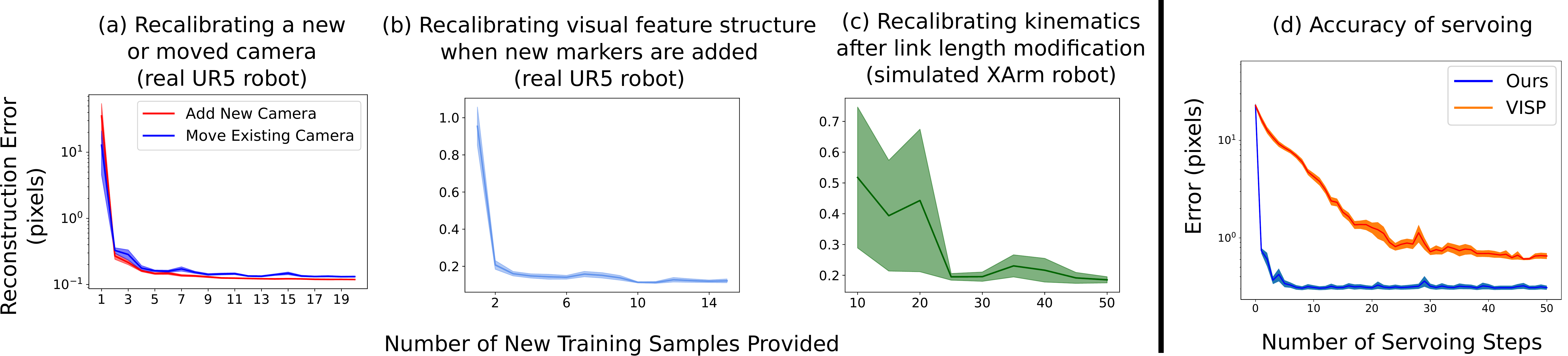}
	\caption{(a,b,c) We evaluate our method's ability to adapt to changes to the robot-camera setup by measuring the model's reconstruction error as it sequentially receives new observations and updates the model parameters online. (a) On the real UR5, we measure the reconstruction error in both the case where we add a new camera and the case where we move an existing camera. (b) On the real UR5, we measure the reconstruction error when we attach new visual features to the end-effector. (c) On a simulated UR5, we measure the reconstruction error when we modify the link lengths of the robot, simulating a change to the robot kinematics. (d) We compare the accuracy of our method with VISP \cite{marchand2005visp} on servoing to randomly select target coordinates of the visual features.}
\label{fig:graphs}
\end{figure*}

Finally, we demonstrate that our architecture enables accurate visual servoing. Figure \ref{fig:graphs}d show a comparison between our method and VISP \cite{marchand2005visp}, a popular library implementing visual servoing. In this evaluation, we first randomly select target locations for the visual features and then allow each method to servo to the target. We measure the error between the target and actual pixel coordinates at each time step. (We average over 50 runs with different randomly selected targets). We found that both DurableVS and VISP achieve sub-pixel accuracy. Our method slightly outperforms VISP, likely due to inaccuracies in camera calibration, which further motivates the value of learning as opposed to assuming a fixed camera calibration\footnote{We used a standard camera calibration routine provided in ROS \cite{quigley2009ros} and validated its accuracy}.
In these experiments, it appears DurableVS converges to the targets more rapidly than VISP. However, we do not claim that DurableVS is faster than VISP in general, as VISP’s convergence rate depends on many factors. We provide this data to show that DurableVS can be competitive with VISP (out of the box) on a real robot, in addition to requiring no calibration and automatically recovering from a broad class of calibration errors.



\section{Related Work}

\subsection{Kinematic \& Camera Calibration}

Robotic systems often assume the robot's kinematic model and any cameras it uses are calibrated. Kinematic calibration often needs precisely manufactured devices or manual alignment procedures \cite{hollerbach1996calibration,renders1991kinematic,ikits1997kinematic} which can be time-consuming.
Camera calibration needs a pattern to be placed at different locations in view of the camera \cite{zhang2000flexible, quigley2009ros,bradski2000opencv, tsai1980review} and this process must be repeated whenever the camera is moved. Other prior works propose methods for learning kinematics \cite{rolf2010goal, atkeson1989learning, d2001learning, ikits1997kinematic, renders1991kinematic} or self-calibration of cameras \cite{hemayed2003survey, faugeras1992camera, fraser1997digital, lenz1989calibrating, tsai1989new}. However, none of these methods can learn both kinematic and camera models, unsupervised.

Independent kinematic and camera calibration can also negatively affect accuracy of visual servoing since errors in the kinematic calibration can lead to errors in camera calibration. Many prior works explore the benefits of simultaneous kinematics and camera calibration \cite{zhuang1995simultaneous, kummerle2012simultaneous, tsai1989new} in avoiding these compounded errors. However, these calibration approaches are aimed at identifying the veridical system parameters and thus cannot be learned without supervision.

Finally, kinematics and camera calibration must be repeated when even small changes are made to the robot-camera system. As demonstrated in our experiments, our approach can quickly adapt to such changes online, enabling the system to continue to function without interruption or intervention.

\subsection{Visual Servoing} 

Visual servoing refers to techniques that use visual feedback to guide control. Commonly used methods for visual servoing are image-based visual servoing (IBVS) and position-based visual servoing (PBVS). In general, these methods seek to move certain tracked image features to desired target locations. Most proposed visual-servoing approaches depend on calibration of the robot's kinematic model and cameras \cite{mohta2014vision,espiau1992new,mohta2014vision, chaumette2016visual, siciliano2016springer}. To relax this assumption some methods directly estimate the image-joint Jacobian, capturing how changes to the robot's joint positions lead to changes in image features \cite{jagersand1997experimental, massoud2007global, hosoda1998adaptive, piepmeier2004uncalibrated, yoshimi1994active}. Unlike differential approaches, our method learns a structured model of the system. 

Visual Servoing Platform (VISP) \cite{marchand2005visp} is a popular library implementing various visual servoing control laws. Part of our contribution in this work is the Durable Visual Servoing (DVS) Python \cite{van1995python} package which implements our architecture in PyTorch \cite{pytorch}. This package includes a demo of data-collection, learning, and visual servoing with the learned model in a simulated PyBullet \cite{coumans2017pybullet} environment. In addition, we provide tutorials to enable researchers to apply our system to their applications and to build upon our architecture.

\subsection{Learning Vision-based Control}

Recent work on learning control of robots using visual feedback has focused on deep learning approaches. \cite{sadeghi2018sim2real} learns viewpoint-invariant visual servoing but requires pre-training in simulated environments before transferring to real environments. \cite{levine2016end} trains perception and control systems simultaneously using an end-to-end neural network and demonstrates that this can be applied to perform a wide range of tasks. However, as is often the case with neural networks, they demand a large amount of training data and time. With our method, models can be learned from significantly fewer samples and less time than these neural methods. 

In future work, we plan to explore the combination of our structured generative model with deep learning approaches. We believe that this combination can improve the overall data-efficiency of end-to-end learning of robotic tasks like grasping \cite{levine2018learning, pinto2016supersizing}, drone flying \cite{gandhi2017learning}, and object manipulation/interaction \cite{agrawal2016learning}. Consider if the deep learned system could robustly detect visual features and plan the path of those image features such that the robot performs the desired task. Our architecture could then take this as input and infer the necessary joint commands to perform the task. In this paper, we used Aruco markers \cite{garrido2014automatic} attached to the end-effector as the visual features. By combining our architecture with deep learning, we could develop an overall more general system.

%

\section{Conclusion}

In this paper, we presented an approach to learning visual servoing based on structured generative models. We have shown that given an arbitrary uncalibrated robot observed through an arbitrary number of uncalibrated cameras, our system can learn accurate visual servoing. This is done completely unsupervised and extremely data-efficiently (in less than a minute). Learning can continue online, enabling the system to detect and recover from changes to the robot-camera system. The problem we have addressed in this work is broadly applicable to many vision-based robotics domains and tasks. We hope that our contribution can enable more data-efficient and robust systems that can operate in real-world settings with no human supervision.

\bibliography{rvs}

\begin{thebibliography}{36}
\providecommand{\natexlab}[1]{#1}
\providecommand{\url}[1]{\texttt{#1}}
\expandafter\ifx\csname urlstyle\endcsname\relax
  \providecommand{\doi}[1]{doi: #1}\else
  \providecommand{\doi}{doi: \begingroup \urlstyle{rm}\Url}\fi

\bibitem[Wolpert et~al.(1995)Wolpert, Ghahramani, and
  Jordan]{wolpert1995internal}
D.~M. Wolpert, Z.~Ghahramani, and M.~I. Jordan.
\newblock An internal model for sensorimotor integration.
\newblock \emph{Science}, 269\penalty0 (5232):\penalty0 1880--1882, 1995.

\bibitem[Attinger et~al.(2017)Attinger, Wang, and
  Keller]{attinger2017visuomotor}
A.~Attinger, B.~Wang, and G.~B. Keller.
\newblock Visuomotor coupling shapes the functional development of mouse visual
  cortex.
\newblock \emph{Cell}, 169\penalty0 (7):\penalty0 1291--1302, 2017.

\bibitem[Kragic and Christensen(2002)]{kragic2002survey}
D.~Kragic and H.~I. Christensen.
\newblock Survey on visual servoing for manipulation.
\newblock In \emph{USENIX Technical Conference}, 2002.

\bibitem[Chaumette et~al.(2016)Chaumette, Hutchinson, and
  Corke]{chaumette2016visual}
F.~Chaumette, S.~Hutchinson, and P.~Corke.
\newblock Visual servoing.
\newblock In \emph{Springer Handbook of Robotics}, pages 841--866. Springer,
  2016.

\bibitem[Hollerbach and Wampler(1996)]{hollerbach1996calibration}
J.~M. Hollerbach and C.~W. Wampler.
\newblock The calibration index and taxonomy for robot kinematic calibration
  methods.
\newblock \emph{The international journal of robotics research}, 15\penalty0
  (6):\penalty0 573--591, 1996.

\bibitem[Renders et~al.(1991)Renders, Rossignol, Becquet, Hanus,
  et~al.]{renders1991kinematic}
J.-M. Renders, E.~Rossignol, M.~Becquet, R.~Hanus, et~al.
\newblock Kinematic calibration and geometrical parameter identification for
  robots.
\newblock \emph{IEEE Transactions on robotics and automation}, 7\penalty0
  (6):\penalty0 721--732, 1991.

\bibitem[Ikits and Hollerbach(1997)]{ikits1997kinematic}
M.~Ikits and J.~M. Hollerbach.
\newblock Kinematic calibration using a plane constraint.
\newblock In \emph{Proceedings of International Conference on Robotics and
  Automation}, volume~4, pages 3191--3196. IEEE, 1997.

\bibitem[Zhang(2000)]{zhang2000flexible}
Z.~Zhang.
\newblock A flexible new technique for camera calibration.
\newblock \emph{IEEE Transactions on pattern analysis and machine
  intelligence}, 22, 2000.

\bibitem[Hemayed(2003)]{hemayed2003survey}
E.~Hemayed.
\newblock A survey of camera self-calibration.
\newblock In \emph{Proceedings of the IEEE Conference on Advanced Video and
  Signal Based Surveillance, 2003.}, pages 351--357. IEEE, 2003.

\bibitem[Quigley et~al.(2009)Quigley, Conley, Gerkey, Faust, Foote, Leibs,
  Wheeler, and Ng]{quigley2009ros}
M.~Quigley, K.~Conley, B.~Gerkey, J.~Faust, T.~Foote, J.~Leibs, R.~Wheeler, and
  A.~Y. Ng.
\newblock Ros: an open-source robot operating system.
\newblock In \emph{ICRA workshop on open source software}, volume~3, page~5.
  Kobe, Japan, 2009.

\bibitem[Bradski and Kaehler(2000)]{bradski2000opencv}
G.~Bradski and A.~Kaehler.
\newblock Opencv.
\newblock \emph{Dr. Dobb’s journal of software tools}, 3, 2000.

\bibitem[Zhuang et~al.(1995)Zhuang, Wang, and Roth]{zhuang1995simultaneous}
H.~Zhuang, K.~Wang, and Z.~S. Roth.
\newblock Simultaneous calibration of a robot and a hand-mounted camera.
\newblock \emph{IEEE Transactions on Robotics and Automation}, 11\penalty0
  (5):\penalty0 649--660, 1995.

\bibitem[K{\"u}mmerle et~al.(2012)K{\"u}mmerle, Grisetti, and
  Burgard]{kummerle2012simultaneous}
R.~K{\"u}mmerle, G.~Grisetti, and W.~Burgard.
\newblock Simultaneous parameter calibration, localization, and mapping.
\newblock \emph{Advanced Robotics}, 26\penalty0 (17):\penalty0 2021--2041,
  2012.

\bibitem[Mohta et~al.(2014)Mohta, Kumar, and Daniilidis]{mohta2014vision}
K.~Mohta, V.~Kumar, and K.~Daniilidis.
\newblock Vision-based control of a quadrotor for perching on lines.
\newblock In \emph{2014 IEEE International Conference on Robotics and
  Automation (ICRA)}, pages 3130--3136. IEEE, 2014.

\bibitem[Espiau et~al.(1992)Espiau, Chaumette, and Rives]{espiau1992new}
B.~Espiau, F.~Chaumette, and P.~Rives.
\newblock A new approach to visual servoing in robotics.
\newblock \emph{ieee Transactions on Robotics and Automation}, 8\penalty0
  (3):\penalty0 313--326, 1992.

\bibitem[Jagersand et~al.(1997)Jagersand, Fuentes, and
  Nelson]{jagersand1997experimental}
M.~Jagersand, O.~Fuentes, and R.~Nelson.
\newblock Experimental evaluation of uncalibrated visual servoing for precision
  manipulation.
\newblock In \emph{Proceedings of International Conference on Robotics and
  Automation}, volume~4, pages 2874--2880. IEEE, 1997.

\bibitem[massoud Farahmand et~al.(2007)massoud Farahmand, Shademan, and
  Jagersand]{massoud2007global}
A.~massoud Farahmand, A.~Shademan, and M.~Jagersand.
\newblock Global visual-motor estimation for uncalibrated visual servoing.
\newblock In \emph{2007 IEEE/RSJ International Conference on Intelligent Robots
  and Systems}, pages 1969--1974. IEEE, 2007.

\bibitem[Hosoda et~al.(1998)Hosoda, Igarashi, and Asada]{hosoda1998adaptive}
K.~Hosoda, K.~Igarashi, and M.~Asada.
\newblock Adaptive hybrid control for visual and force servoing in an unknown
  environment.
\newblock \emph{IEEE Robotics \& Automation Magazine}, 5\penalty0 (4):\penalty0
  39--43, 1998.

\bibitem[Piepmeier et~al.(2004)Piepmeier, McMurray, and
  Lipkin]{piepmeier2004uncalibrated}
J.~A. Piepmeier, G.~V. McMurray, and H.~Lipkin.
\newblock Uncalibrated dynamic visual servoing.
\newblock \emph{IEEE Transactions on Robotics and Automation}, 20\penalty0
  (1):\penalty0 143--147, 2004.

\bibitem[Sadeghi et~al.(2018)Sadeghi, Toshev, Jang, and
  Levine]{sadeghi2018sim2real}
F.~Sadeghi, A.~Toshev, E.~Jang, and S.~Levine.
\newblock Sim2real viewpoint invariant visual servoing by recurrent control.
\newblock In \emph{Proceedings of the IEEE Conference on Computer Vision and
  Pattern Recognition}, pages 4691--4699, 2018.

\bibitem[Levine et~al.(2016)Levine, Finn, Darrell, and Abbeel]{levine2016end}
S.~Levine, C.~Finn, T.~Darrell, and P.~Abbeel.
\newblock End-to-end training of deep visuomotor policies.
\newblock \emph{The Journal of Machine Learning Research}, 17\penalty0
  (1):\penalty0 1334--1373, 2016.

\bibitem[Levine et~al.(2018)Levine, Pastor, Krizhevsky, Ibarz, and
  Quillen]{levine2018learning}
S.~Levine, P.~Pastor, A.~Krizhevsky, J.~Ibarz, and D.~Quillen.
\newblock Learning hand-eye coordination for robotic grasping with deep
  learning and large-scale data collection.
\newblock \emph{The International Journal of Robotics Research}, 37\penalty0
  (4-5):\penalty0 421--436, 2018.

\bibitem[Pinto and Gupta(2016)]{pinto2016supersizing}
L.~Pinto and A.~Gupta.
\newblock Supersizing self-supervision: Learning to grasp from 50k tries and
  700 robot hours.
\newblock In \emph{2016 IEEE international conference on robotics and
  automation (ICRA)}, pages 3406--3413. IEEE, 2016.

\bibitem[Gandhi et~al.(2017)Gandhi, Pinto, and Gupta]{gandhi2017learning}
D.~Gandhi, L.~Pinto, and A.~Gupta.
\newblock Learning to fly by crashing.
\newblock In \emph{2017 IEEE/RSJ International Conference on Intelligent Robots
  and Systems (IROS)}, pages 3948--3955. IEEE, 2017.

\bibitem[Agrawal et~al.(2016)Agrawal, Nair, Abbeel, Malik, and
  Levine]{agrawal2016learning}
P.~Agrawal, A.~V. Nair, P.~Abbeel, J.~Malik, and S.~Levine.
\newblock Learning to poke by poking: Experiential learning of intuitive
  physics.
\newblock In \emph{Advances in neural information processing systems}, pages
  5074--5082, 2016.

\bibitem[Hartenberg and Danavit(1964)]{hartenberg1964kinematic}
R.~Hartenberg and J.~Danavit.
\newblock \emph{Kinematic synthesis of linkages}.
\newblock New York: McGraw-Hill, 1964.

\bibitem[Szeliski(2010)]{szeliski2010computer}
R.~Szeliski.
\newblock \emph{Computer vision: algorithms and applications}.
\newblock Springer Science \& Business Media, 2010.

\bibitem[Liu and Nocedal(1989)]{liu1989limited}
D.~C. Liu and J.~Nocedal.
\newblock On the limited memory bfgs method for large scale optimization.
\newblock \emph{Mathematical programming}, 45\penalty0 (1-3):\penalty0
  503--528, 1989.

\bibitem[Lepetit et~al.(2009)Lepetit, Moreno-Noguer, and Fua]{lepetit2009epnp}
V.~Lepetit, F.~Moreno-Noguer, and P.~Fua.
\newblock Epnp: An accurate o (n) solution to the pnp problem.
\newblock \emph{International journal of computer vision}, 81\penalty0
  (2):\penalty0 155, 2009.

\bibitem[Siciliano and Khatib(2016)]{siciliano2016springer}
B.~Siciliano and O.~Khatib.
\newblock \emph{Springer handbook of robotics}.
\newblock springer, 2016.

\bibitem[Garrido-Jurado et~al.(2014)Garrido-Jurado, Mu{\~n}oz-Salinas,
  Madrid-Cuevas, and Mar{\'\i}n-Jim{\'e}nez]{garrido2014automatic}
S.~Garrido-Jurado, R.~Mu{\~n}oz-Salinas, F.~J. Madrid-Cuevas, and M.~J.
  Mar{\'\i}n-Jim{\'e}nez.
\newblock Automatic generation and detection of highly reliable fiducial
  markers under occlusion.
\newblock \emph{Pattern Recognition}, 47\penalty0 (6):\penalty0 2280--2292,
  2014.

\bibitem[Marchand et~al.(2005)Marchand, Spindler, and
  Chaumette]{marchand2005visp}
{\'E}.~Marchand, F.~Spindler, and F.~Chaumette.
\newblock Visp for visual servoing: a generic software platform with a wide
  class of robot control skills.
\newblock \emph{IEEE Robotics \& Automation Magazine}, 12\penalty0
  (4):\penalty0 40--52, 2005.

\bibitem[Cremer et~al.(2016)Cremer, Mastromoro, and
  Popa]{cremer2016performance}
S.~Cremer, L.~Mastromoro, and D.~O. Popa.
\newblock On the performance of the baxter research robot.
\newblock In \emph{2016 IEEE international symposium on assembly and
  manufacturing (ISAM)}, pages 106--111. IEEE, 2016.

\bibitem[Van~Rossum and Drake~Jr(1995)]{van1995python}
G.~Van~Rossum and F.~L. Drake~Jr.
\newblock \emph{Python tutorial}.
\newblock Centrum voor Wiskunde en Informatica Amsterdam, The Netherlands,
  1995.

\bibitem[Paszke et~al.(2019)Paszke, Gross, Massa, Lerer, Bradbury, Chanan,
  Killeen, Lin, Gimelshein, Antiga, Desmaison, Kopf, Yang, DeVito, Raison,
  Tejani, Chilamkurthy, Steiner, Fang, Bai, and Chintala]{pytorch}
A.~Paszke, S.~Gross, F.~Massa, A.~Lerer, J.~Bradbury, G.~Chanan, T.~Killeen,
  Z.~Lin, N.~Gimelshein, L.~Antiga, A.~Desmaison, A.~Kopf, E.~Yang, Z.~DeVito,
  M.~Raison, A.~Tejani, S.~Chilamkurthy, B.~Steiner, L.~Fang, J.~Bai, and
  S.~Chintala.
\newblock Pytorch: An imperative style, high-performance deep learning library.
\newblock In H.~Wallach, H.~Larochelle, A.~Beygelzimer, F.~d\textquotesingle
  Alch\'{e}-Buc, E.~Fox, and R.~Garnett, editors, \emph{Advances in Neural
  Information Processing Systems 32}, pages 8024--8035. Curran Associates,
  Inc., 2019.
\newblock URL
  \url{http://papers.neurips.cc/paper/9015-pytorch-an-imperative-style-high-performance-deep-learning-library.pdf}.

\bibitem[Coumans and Bai(2016--2021)]{coumans2021pybullet}
E.~Coumans and Y.~Bai.
\newblock Pybullet, a python module for physics simulation for games, robotics
  and machine learning.
\newblock \url{http://pybullet.org}, 2016--2021.

\end{thebibliography}


\begin{thebibliography}{10}

\bibitem{agrawal2016learning}
Pulkit Agrawal, Ashvin~V Nair, Pieter Abbeel, Jitendra Malik, and Sergey
  Levine.
\newblock Learning to poke by poking: Experiential learning of intuitive
  physics.
\newblock In {\em Advances in neural information processing systems}, pages
  5074--5082, 2016.

\bibitem{atkeson1989learning}
Christopher~G Atkeson.
\newblock Learning arm kinematics and dynamics.
\newblock {\em Annual review of neuroscience}, 12(1):157--183, 1989.

\bibitem{beeson2015trac}
Patrick Beeson and Barrett Ames.
\newblock Trac-ik: An open-source library for improved solving of generic
  inverse kinematics.
\newblock In {\em 2015 IEEE-RAS 15th International Conference on Humanoid
  Robots (Humanoids)}, pages 928--935. IEEE, 2015.

\bibitem{bradski2000opencv}
Gary Bradski and Adrian Kaehler.
\newblock Opencv.
\newblock {\em Dr. Dobb’s journal of software tools}, 3, 2000.

\bibitem{chaumette2016visual}
Fran{\c{c}}ois Chaumette, Seth Hutchinson, and Peter Corke.
\newblock Visual servoing.
\newblock In {\em Springer Handbook of Robotics}, pages 841--866. Springer,
  2016.

\bibitem{coumans2017pybullet}
Erwin Coumans and Yunfei Bai.
\newblock Pybullet, a python module for physics simulation in robotics, games
  and machine learning, 2017.

\bibitem{d2001learning}
Aaron D'Souza, Sethu Vijayakumar, and Stefan Schaal.
\newblock Learning inverse kinematics.
\newblock In {\em Proceedings 2001 IEEE/RSJ International Conference on
  Intelligent Robots and Systems. Expanding the Societal Role of Robotics in
  the the Next Millennium (Cat. No. 01CH37180)}, volume~1, pages 298--303.
  IEEE, 2001.

\bibitem{espiau1992new}
Bernard Espiau, Fran{\c{c}}ois Chaumette, and Patrick Rives.
\newblock A new approach to visual servoing in robotics.
\newblock {\em ieee Transactions on Robotics and Automation}, 8(3):313--326,
  1992.

\bibitem{faugeras1992camera}
Olivier~D Faugeras, Q-T Luong, and Stephen~J Maybank.
\newblock Camera self-calibration: Theory and experiments.
\newblock In {\em European conference on computer vision}, pages 321--334.
  Springer, 1992.

\bibitem{fischler1981random}
Martin~A Fischler and Robert~C Bolles.
\newblock Random sample consensus: a paradigm for model fitting with
  applications to image analysis and automated cartography.
\newblock {\em Communications of the ACM}, 24(6):381--395, 1981.

\bibitem{fraser1997digital}
Clive~S Fraser.
\newblock Digital camera self-calibration.
\newblock {\em ISPRS Journal of Photogrammetry and Remote sensing},
  52(4):149--159, 1997.

\bibitem{gandhi2017learning}
Dhiraj Gandhi, Lerrel Pinto, and Abhinav Gupta.
\newblock Learning to fly by crashing.
\newblock In {\em 2017 IEEE/RSJ International Conference on Intelligent Robots
  and Systems (IROS)}, pages 3948--3955. IEEE, 2017.

\bibitem{garrido2014automatic}
Sergio Garrido-Jurado, Rafael Mu{\~n}oz-Salinas, Francisco~Jos{\'e}
  Madrid-Cuevas, and Manuel~Jes{\'u}s Mar{\'\i}n-Jim{\'e}nez.
\newblock Automatic generation and detection of highly reliable fiducial
  markers under occlusion.
\newblock {\em Pattern Recognition}, 47(6):2280--2292, 2014.

\bibitem{hartenberg1964kinematic}
Richard Hartenberg and Jacques Danavit.
\newblock {\em Kinematic synthesis of linkages}.
\newblock New York: McGraw-Hill, 1964.

\bibitem{hemayed2003survey}
Elsayed Hemayed.
\newblock A survey of camera self-calibration.
\newblock In {\em Proceedings of the IEEE Conference on Advanced Video and
  Signal Based Surveillance, 2003.}, pages 351--357. IEEE, 2003.

\bibitem{hollerbach1996calibration}
John~M Hollerbach and Charles~W Wampler.
\newblock The calibration index and taxonomy for robot kinematic calibration
  methods.
\newblock {\em The international journal of robotics research}, 15(6):573--591,
  1996.

\bibitem{hosoda1998adaptive}
Koh Hosoda, Katsuji Igarashi, and Minoru Asada.
\newblock Adaptive hybrid control for visual and force servoing in an unknown
  environment.
\newblock {\em IEEE Robotics \& Automation Magazine}, 5(4):39--43, 1998.

\bibitem{ikits1997kinematic}
Milan Ikits and John~M Hollerbach.
\newblock Kinematic calibration using a plane constraint.
\newblock In {\em Proceedings of International Conference on Robotics and
  Automation}, volume~4, pages 3191--3196. IEEE, 1997.

\bibitem{jagersand1997experimental}
Martin Jagersand, Olac Fuentes, and Randal Nelson.
\newblock Experimental evaluation of uncalibrated visual servoing for precision
  manipulation.
\newblock In {\em Proceedings of International Conference on Robotics and
  Automation}, volume~4, pages 2874--2880. IEEE, 1997.

\bibitem{kragic2002survey}
Danica Kragic and Henrik~I Christensen.
\newblock Survey on visual servoing for manipulation.
\newblock In {\em USENIX Technical Conference}, 2002.

\bibitem{kummerle2012simultaneous}
Rainer K{\"u}mmerle, Giorgio Grisetti, and Wolfram Burgard.
\newblock Simultaneous parameter calibration, localization, and mapping.
\newblock {\em Advanced Robotics}, 26(17):2021--2041, 2012.

\bibitem{lenz1989calibrating}
Reimar~K Lenz and Roger~Y Tsai.
\newblock Calibrating a cartesian robot with eye-on-hand configuration
  independent of eye-to-hand relationship.
\newblock {\em IEEE Transactions on Pattern Analysis \& Machine Intelligence},
  (9):916--928, 1989.

\bibitem{levine2016end}
Sergey Levine, Chelsea Finn, Trevor Darrell, and Pieter Abbeel.
\newblock End-to-end training of deep visuomotor policies.
\newblock {\em The Journal of Machine Learning Research}, 17(1):1334--1373,
  2016.

\bibitem{levine2018learning}
Sergey Levine, Peter Pastor, Alex Krizhevsky, Julian Ibarz, and Deirdre
  Quillen.
\newblock Learning hand-eye coordination for robotic grasping with deep
  learning and large-scale data collection.
\newblock {\em The International Journal of Robotics Research},
  37(4-5):421--436, 2018.

\bibitem{liu1989limited}
Dong~C Liu and Jorge Nocedal.
\newblock On the limited memory bfgs method for large scale optimization.
\newblock {\em Mathematical programming}, 45(1-3):503--528, 1989.

\bibitem{marchand2005visp}
{\'E}ric Marchand, Fabien Spindler, and Fran{\c{c}}ois Chaumette.
\newblock Visp for visual servoing: a generic software platform with a wide
  class of robot control skills.
\newblock {\em IEEE Robotics \& Automation Magazine}, 12(4):40--52, 2005.

\bibitem{massoud2007global}
Amir massoud Farahmand, Azad Shademan, and Martin Jagersand.
\newblock Global visual-motor estimation for uncalibrated visual servoing.
\newblock In {\em 2007 IEEE/RSJ International Conference on Intelligent Robots
  and Systems}, pages 1969--1974. IEEE, 2007.

\bibitem{mohta2014vision}
Kartik Mohta, Vijay Kumar, and Kostas Daniilidis.
\newblock Vision-based control of a quadrotor for perching on lines.
\newblock In {\em 2014 IEEE International Conference on Robotics and Automation
  (ICRA)}, pages 3130--3136. IEEE, 2014.

\bibitem{pytorch}
Adam Paszke, Sam Gross, Francisco Massa, Adam Lerer, James Bradbury, Gregory
  Chanan, Trevor Killeen, Zeming Lin, Natalia Gimelshein, Luca Antiga, Alban
  Desmaison, Andreas Kopf, Edward Yang, Zachary DeVito, Martin Raison, Alykhan
  Tejani, Sasank Chilamkurthy, Benoit Steiner, Lu~Fang, Junjie Bai, and Soumith
  Chintala.
\newblock Pytorch: An imperative style, high-performance deep learning library.
\newblock In H.~Wallach, H.~Larochelle, A.~Beygelzimer, F.~d\textquotesingle
  Alch\'{e}-Buc, E.~Fox, and R.~Garnett, editors, {\em Advances in Neural
  Information Processing Systems 32}, pages 8024--8035. Curran Associates,
  Inc., 2019.

\bibitem{piepmeier2004uncalibrated}
Jenelle~Armstrong Piepmeier, Gary~V McMurray, and Harvey Lipkin.
\newblock Uncalibrated dynamic visual servoing.
\newblock {\em IEEE Transactions on Robotics and Automation}, 20(1):143--147,
  2004.

\bibitem{pinto2016supersizing}
Lerrel Pinto and Abhinav Gupta.
\newblock Supersizing self-supervision: Learning to grasp from 50k tries and
  700 robot hours.
\newblock In {\em 2016 IEEE international conference on robotics and automation
  (ICRA)}, pages 3406--3413. IEEE, 2016.

\bibitem{quigley2009ros}
Morgan Quigley, Ken Conley, Brian Gerkey, Josh Faust, Tully Foote, Jeremy
  Leibs, Rob Wheeler, and Andrew~Y Ng.
\newblock Ros: an open-source robot operating system.
\newblock In {\em ICRA workshop on open source software}, volume~3, page~5.
  Kobe, Japan, 2009.

\bibitem{renders1991kinematic}
Jean-Michel Renders, Eric Rossignol, Marc Becquet, Raymond Hanus, et~al.
\newblock Kinematic calibration and geometrical parameter identification for
  robots.
\newblock {\em IEEE Transactions on robotics and automation}, 7(6):721--732,
  1991.

\bibitem{rolf2010goal}
Matthias Rolf, Jochen~J Steil, and Michael Gienger.
\newblock Goal babbling permits direct learning of inverse kinematics.
\newblock {\em IEEE Transactions on Autonomous Mental Development},
  2(3):216--229, 2010.

\bibitem{sadeghi2018sim2real}
Fereshteh Sadeghi, Alexander Toshev, Eric Jang, and Sergey Levine.
\newblock Sim2real viewpoint invariant visual servoing by recurrent control.
\newblock In {\em Proceedings of the IEEE Conference on Computer Vision and
  Pattern Recognition}, pages 4691--4699, 2018.

\bibitem{siciliano2016springer}
Bruno Siciliano and Oussama Khatib.
\newblock {\em Springer handbook of robotics}.
\newblock springer, 2016.

\bibitem{szeliski2010computer}
Richard Szeliski.
\newblock {\em Computer vision: algorithms and applications}.
\newblock Springer Science \& Business Media, 2010.

\bibitem{tsai1980review}
Roger~Y Tsai and Reimar Lenz.
\newblock Review of the two-stage camera calibration technique plus some new
  implementation tips and some new techniques for center and scale calibration.
\newblock In {\em Optical Society of America, Topical Meeting on Machine
  Vision}, 1980.

\bibitem{tsai1989new}
Roger~Y Tsai and Reimar~K Lenz.
\newblock A new technique for fully autonomous and efficient 3d robotics
  hand/eye calibration.
\newblock {\em IEEE Transactions on robotics and automation}, 5(3):345--358,
  1989.

\bibitem{van1995python}
Guido Van~Rossum and Fred~L Drake~Jr.
\newblock {\em Python tutorial}.
\newblock Centrum voor Wiskunde en Informatica Amsterdam, The Netherlands,
  1995.

\bibitem{yoshimi1994active}
Billibon~H Yoshimi and Peter~K Allen.
\newblock Active, uncalibrated visual servoing.
\newblock In {\em Proceedings of the 1994 IEEE International Conference on
  Robotics and Automation}, pages 156--161. IEEE, 1994.

\bibitem{zhang2000flexible}
Zhengyou Zhang.
\newblock A flexible new technique for camera calibration.
\newblock {\em IEEE Transactions on pattern analysis and machine intelligence},
  22, 2000.

\bibitem{zhuang1995simultaneous}
Hanqi Zhuang, Kuanchih Wang, and Zvi~S Roth.
\newblock Simultaneous calibration of a robot and a hand-mounted camera.
\newblock {\em IEEE Transactions on Robotics and Automation}, 11(5):649--660,
  1995.

\end{thebibliography}


\begin{thebibliography}{3}
\providecommand{\natexlab}[1]{#1}
\providecommand{\url}[1]{\texttt{#1}}
\expandafter\ifx\csname urlstyle\endcsname\relax
  \providecommand{\doi}[1]{doi: #1}\else
  \providecommand{\doi}{doi: \begingroup \urlstyle{rm}\Url}\fi

\bibitem[Nist{\'e}r(2004)]{nister2004efficient}
D.~Nist{\'e}r.
\newblock An efficient solution to the five-point relative pose problem.
\newblock \emph{IEEE transactions on pattern analysis and machine
  intelligence}, 26\penalty0 (6):\penalty0 0756--777, 2004.

\bibitem[Arun et~al.(1987)Arun, Huang, and Blostein]{arun1987least}
K.~S. Arun, T.~S. Huang, and S.~D. Blostein.
\newblock Least-squares fitting of two 3-d point sets.
\newblock \emph{IEEE Transactions on pattern analysis and machine
  intelligence}, \penalty0 (5):\penalty0 698--700, 1987.

\bibitem[Schonberger and Frahm(2016)]{schonberger2016structure}
J.~L. Schonberger and J.-M. Frahm.
\newblock Structure-from-motion revisited.
\newblock In \emph{Proceedings of the IEEE Conference on Computer Vision and
  Pattern Recognition}, pages 4104--4113, 2016.

\end{thebibliography}
\bibliographystyle{plain}

\end{document}


\maketitle

\appendix

\section{Initialization of Camera and Feature Coordinate Learning}

In the main text, we noted that the optimization of camera parameters $\mathbf{V}_{1:c}$, feature coordinate parameters $\mathbf{F}$, and end-effector poses $\mathbf{p}_{1:T}$, given the image feature detections $\mathbf{I}_t$ depends on selecting a good initialization since the optimization is prone converging to sub-optimal local minima. In the following subsections we describe two methods for computing this initialization. The first is more robust but depends on observing the robot in 2 cameras. The second can be applied when there is only 1 camera.

\subsection{Initialization by Triangulation}
If the cameras are placed such that in many timesteps features are observed in multiple cameras, we can leverage this to compute an initialization for our optimization. With 5 feature correspondences between cameras, we can estimate the fundamental matrix between the cameras \citep{nister2004efficient}. This means we only need 5 timesteps at which a feature was observed in both cameras. From the fundamental matrix and an estimate of the intrinsic parameters (we use the factory-provided intrinsic estimate) we can get the essential matrix. Then from this essential matrix, we can recover 4 possible solutions for the camera baseline, the transformation between the two cameras. To select one of these 4 solutions, we triangulate the observed points to 3D and select the configuration of the cameras that places the most observed features in front of both cameras.

This procedure can be repeated for pairs of cameras that observe 5 of the same features until we have accounted for all the cameras. If a camera cannot be added this procedure (if it does not observe sufficiently many features) then we can ignore it from the initial optimization and reincorporate it later in the learning procedure.

With this estimate of the poses of the cameras, we can triangulate all the features observed in multiple cameras to 3D. Then at each timestep, we have the 3D coordinates of those features. Now, between two timesteps for which there are at least 3 of the same features triangulated, we can estimate the optimal (in terms of least square error) rigid transformation between the feature coordinates at the timesteps \citep{arun1987least}. This optimal transformation can be found by taking the vector between the centroids of the coordinates as the translation and then performing an SVD to recover the rotation. We repeat this for pairs of timesteps that share at least 3 triangulated features until we have accounted for all timesteps. If we are unable to add a timestep because it does not share sufficiently many triangulated features with another timestep, we ignore it.

Since the features we track are rigidly attached to the end-effector, the transformation of the end-effector between timesteps is exactly the transformation of these 3D feature coordinates between timesteps, which we can estimate in closed form. We can compose these transformation across timesteps to recover an estimate of the end-effector Cartesian pose at each timestep.

This procedure yields estimates of (a) the poses of the cameras; (b) the pose of the end-effector at each timestep. Though triangulation in general can be inaccurate, we have found that it provides a sufficiently good initialization for the learning.

\subsection{Initialization by Structure from Motion}\label{sec:sfm}
Since the Triangulation method depends on having at least two cameras, when only a single camera is available, we fall back to using a Structure from Motion procedure. While, most often, Structure from Motion is applied to the case of a fixed object and a moving camera, the same method can be applied to the case of a fixed camera and a moving object (the robot arm). We apply an incremental Structure from Motion procedure that chooses a pair of images with feature correspondences, estimates the camera baseline, and triangulates the features to 3D \citep{schonberger2016structure}. Then we incrementally add information from images that observe features that have already been triangulated and improve the estimation of the object's structure by re-triangulating.

This procedure yields a) the structure of the object; b) the poses of the centroid of that object at each timestep, which then lets us find the rigid transform of the end-effector pose; and c) the poses of the cameras. Again, while this procedure may have inaccuracies, it provides a sufficiently good initialization for the learning, in practice. As is described in a later section, this procedure is quite sensitive to the two initial views that are selected.

\section{Learning from Joint Deltas}
\label{sec:unobserved}
When the joint positions are not observed, we can rely on only the actions (joint position deltas $\mathbf{a}_t = \mathbf{j}_{t+1} - \mathbf{j}_t$) and image features to learn our model. Thus the joint positions at each timestep are also a free parameter and must be optimized along with all the model parameters. The objective function for this optimization is equivalent to that of the end-to-end optimization excluding an additional term. This additional term penalizes the error between the commanded actions and differences of consecutive joint positions, since in the case a noiseless controller these should be equal. The $\lambda$ coefficient, which must be selected, should therefore scale inversely proportionally with the amount of noise in the controller. Note that as $\lambda \xrightarrow{} \infty$, the differences between consecutive positions will be forced to be equal to the actions. Then, we recover the exact objective function of end-to-end optimization.

\begin{equation}
\label{eq:losse2e}
\begin{gathered}
\mathbf{j}_{1:T}^*, \mathbf{R}^f, \mathbf{F}^f, \mathbf{V}_{1:c}^f = \\ \underset{\mathbf{j}_{1:T}, \mathbf{R}, \mathbf{F}, \mathbf{V}_{1:c}}{\text{argmin}} \sum_{t=1}^{T} \sum_{i=1}^{c} D_{\text{pixel}}(C_{\mathbf{V}_i}(S_\mathbf{F}(K_{\mathbf{R}}(\mathbf{j}_t))),\mathbf{I}_{t}^{(i)}) + \lambda \sum_{t=1}^{T-1} \  \|(\mathbf{j}_{t+1} - \mathbf{j}_t, \mathbf{a}_t)\|_2^2
\end{gathered}
\end{equation}

To ensure convergence of this optimization, we first assume that the commanded actions were executed perfectly and run all the optimizations described in previous sections. This is equivalent to optimizing with noisy estimates of the joint positions. This estimate becomes increasingly noisy with each sample, such that after the k\textsuperscript{th} point (for some value of k), the real and estimated position are uncorrelated. Thus, this estimate is only valid for a small number of samples. We use the results of the previous optimizations as an initialization for this optimization, which can correct the noisy joint position while still regularized by the penalty term.

\subsection{Disambiguating Kinematic Redundancy}

In this setting of only observing joint deltas the true joint positions are unknown. Since the function from joint position to end-effector pose is non-injective, there may exist end-effector poses for which there are multiple joint positions that would place the end-effector at that pose. Inference in our model could converge to both possible explanations and would be unable to disambiguate between those explanations. 

In order to resolve this ambiguity, we use inferences over consecutive timesteps. Consider the following situation, where at time $t=1$, we infer two possible joint positions $\mathbf{j}_1^{(1)}$ and $\mathbf{j}_1^{(2)}$. We then command the robot to move by a delta of some $\mathbf{a}$ (so if the current true joint positions are $\mathbf{j}^*$ the robot would move to $\mathbf{j}^* + \mathbf{a}$). Now, at $t=2$ we infer the joint positions to be $\mathbf{j}_2^{(1)}$ and $\mathbf{j}_2^{(2)}$. We can compute the pairwise differences between joint position inferences at $t=1$ and $t=2$ to find which is approximately $\mathbf{a}$. Repeating this over the course of many timesteps will allow us to disambiguate between the joint position explanations.

\section{Demonstrations}

For grasping, we first learn the structure of the target object by placing it in the gripper and collecting samples from 5 views. Now, when the object is placed at a random location, we can compute the joint positions corresponding to the object being within the gripper. Since we learned the structure parameters with the object in the gripper, we do this inference considering only the object's features. Then, we servo using the end-effector features to match that target position. To extend this procedure to perform an insertion of the gripped object, we learn the structure of the target insertion location. We can then infer the joint positions corresponding to the gripped object being at that position (in the target). For both grasping and insertion, moving directly to the target position might cause the gripper to collide with the object or insertion target. Instead, we want to perform a gradual approach from above. We do this by modifying the structure parameters and artificially increasing the distance between the object and end-effector features. As we reduce this distance, we can gradually move the gripper down into the correct configuration.

\section{Experimental Details}

All experiments were run a 2.40GHz Intel i9 processor with 32GB RAM and a Nvidia GeForce GTX 1650 Mobile GPU.

\bibliography{rvs}